\documentclass[letterpaper, 10 pt, conference]{ieeeconf}
\usepackage{times}
\usepackage[pdftex]{graphicx}
\usepackage{subfigure}
\usepackage{amsmath,amssymb,amsopn,amstext,amsfonts}
\usepackage{cancel}
\usepackage[space]{cite}
\usepackage{pdfsync}
\usepackage{balance}
\usepackage{color}
\usepackage{mathtools}
\usepackage{algpseudocode}
\usepackage[ruled,vlined,linesnumbered]{algorithm2e}

\algnewcommand\algorithmicforeach{\textbf{for each}}
\algdef{S}[FOR]{ForEach}[1]{\algorithmicforeach\ #1\ \algorithmicdo}

\usepackage{bm}

\usepackage{diagbox}
\usepackage{epstopdf}
\usepackage{pifont}
\usepackage{fixltx2e}
\usepackage{amsmath}
\usepackage{multirow}
\usepackage{url}
\usepackage{verbatim}
\usepackage{footmisc}
\usepackage[linkcolor=black,citecolor=black,urlcolor=black,colorlinks=true]{hyperref}
\usepackage{subfigure}
\usepackage{color}
\usepackage[skip=3pt, font={small}]{caption} 
\usepackage{tabularx}
\usepackage{float}
\newcommand{\todo}[1]{\textcolor{red}{\emph{\bf#1}}}

\graphicspath{{figures/}}
\DeclareGraphicsExtensions{.png,.jpg,.eps,.pdf}
\IEEEoverridecommandlockouts
\title{\LARGE \bf
	A decentralized framework for simultaneous calibration, localization and mapping with multiple LiDARs
}

\author{Jiarong Lin, Xiyuan Liu and Fu Zhang 
	\thanks{J. Lin, X. Liu and F. Zhang are with the Department of Mechanical Engineering, Hong Kong University, Hong Kong SAR., China. {\tt\small $\{$jiarong.lin, xliuba,  fuzhang$\}$@hku.hk}
	}
}%

\begin{document}
	\maketitle

	\newcommand{\note}[1]{\textcolor{red}{\emph{\bf#1}}}
	\newcommand\footnoteref[1]{\protected@xdef\@thefnmark{\ref{#1}}\@footnotemark}
	\newlength{\bibitemsep}\setlength{\bibitemsep}{.0238\baselineskip}
	\newlength{\bibparskip}\setlength{\bibparskip}{0pt}
	\let\oldthebibliography\thebibliography
	\renewcommand\thebibliography[1]{%
		\oldthebibliography{#1}%
		\setlength{\parskip}{\bibitemsep}%
		\setlength{\itemsep}{\bibparskip}%
	}
	\begin{abstract}
		
		LiDAR is playing a more and more essential role in autonomous driving vehicles for objection detection, self localization and mapping. A single LiDAR frequently suffers from hardware failure (e.g., temporary loss of connection) due to the harsh vehicle environment (e.g., temperature, vibration, etc.), or performance degradation due to the lack of sufficient geometry features, especially for solid-state LiDARs with small field of view (FoV). To improve the system robustness and performance in self-localization and mapping, we develop a decentralized framework for simultaneous calibration, localization and mapping with multiple LiDARs. Our proposed framework is based on an extended Kalman filter (EKF), but is specially formulated for decentralized implementation. Such an implementation could potentially distribute the intensive computation among smaller computing devices or resources dedicated for each LiDAR and remove the single point of failure problem.  Then this decentralized formulation is implemented on an unmanned ground vehicle (UGV) carrying 5 low-cost LiDARs and moving at $1.3m/s$ in urban environments. Experiment results show that the proposed method can successfully and simultaneously estimate the vehicle state (i.e., pose and velocity) and all LiDAR extrinsic parameters. The localization accuracy is up to $\textbf{0.2}\%$ on the two datasets we collected. To share our findings and to make contributions to the community, meanwhile enable the readers to verify our work, we will release all our source codes\footnote{\url{https://github.com/hku-mars/decentralized_loam}\label{decenterlized_loam}} and hardware design blueprint\footnote{\url{https://github.com/hku-mars/lidar_car_platfrom}} on our Github.
		
	\end{abstract}
	
	\section{Introduction}\label{sect_intro}
	
	With the ability of localizing positions and constructing local maps, simultaneous locomotion and mapping (SLAM) using sensors like camera, IMU, LiDAR, etc., are serving as the pillars for missions in autonomous driving~\cite{dissanayake2001a}, field survey~\cite{farina2006permanent} and 3D reconstruction~\cite{engel2014lsd}. Though visual SLAM has been widely applied in exploration and navigation tasks~\cite{mur2015orb,kerl2013dense}, LiDAR SLAM~\cite{hess2016real,pierzchala2018,zhang2014loam} is still of significant essence. Compared with visual sensor, LiDAR is capable of providing high frequency 6 DoF state estimation with low-drift and simultaneously yielding a high resolution environment map. Furthermore, LiDAR is more robust to environments with illumination variations, poor light conditions or few optical textures~\cite{taketomi2017visual}.
	
\begin{figure}[t]
		\includegraphics[width=1.0\columnwidth]{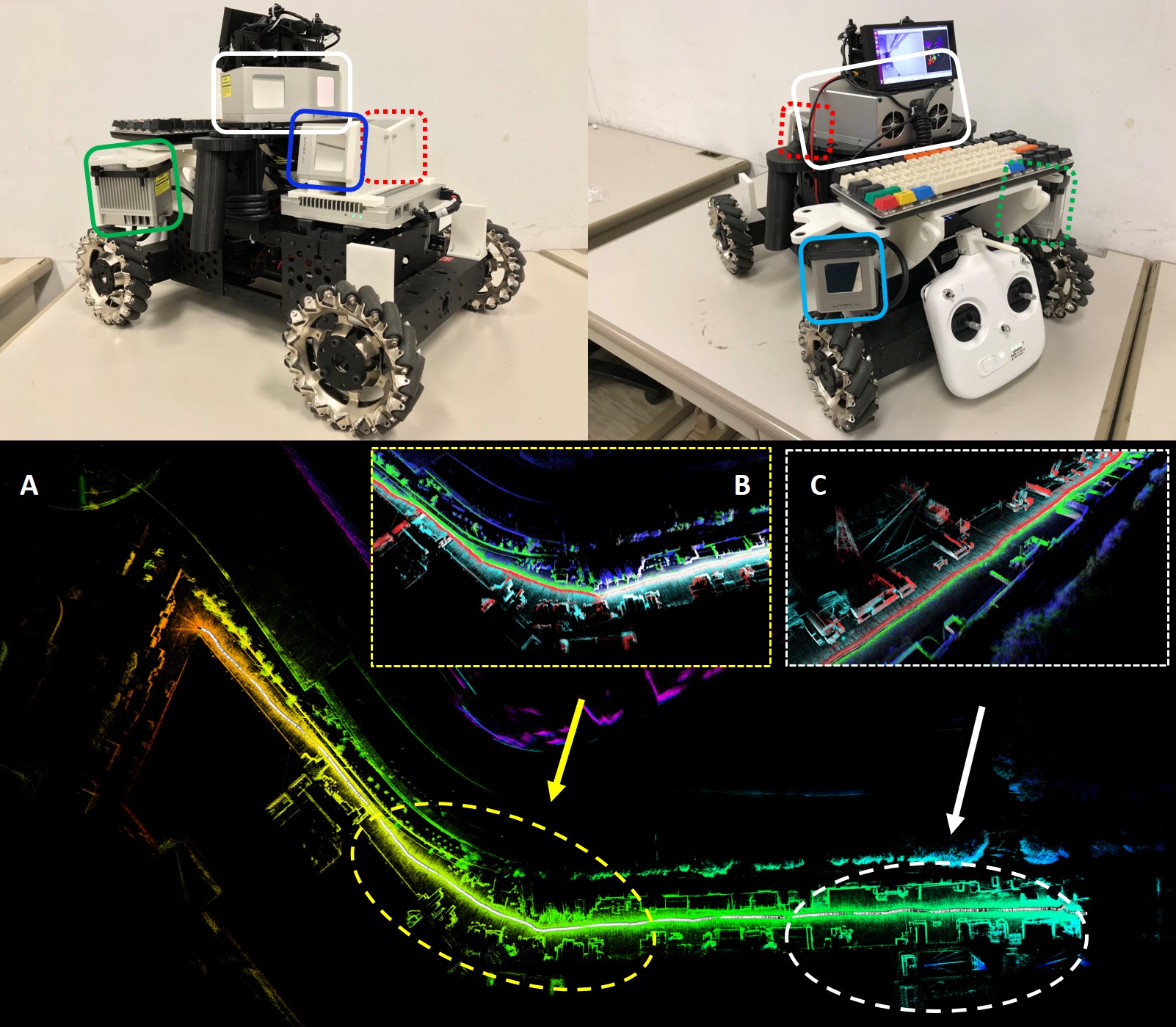}
		\caption{Top: Our decentralized multi-LiDAR vehicle platform. The color of the bounding box is in accordance with the point cloud produced from the same LiDAR (as shown in B, C below). Bottom: A): The map built from $Scene$-1. The point cloud color is calculated by reflectivity. B), C): The detailed point cloud at the start and corner of the map. The color indicates the origin of the LiDAR.}
		\label{fig_lidar_configuration_1}
		\vspace{-1.0cm}
	\end{figure}

	Driven by these widespread robotic applications ~\cite{gao2019flying,xuexi2019slam}, LiDARs have undergone unprecedented developments. In particular, solid state LiDARs have received the most interests~\cite{lin2019,Bosse2012,lin2019fast}. Compared with conventional multi-line spining LiDARs, solid state LiDARs are more cost effective while retaining similar level of performance (e.g., map accuracy, point density). Nevertheless, a major drawback is their small FoV that they are prone to degeneration when facing geometrical feature-less scenes (e.g., wall, sky or grassland). To overcome this, multiple LiDARs are usually embedded at different locations of the vehicle and communicate via the vehicle bus (e.g., controller area network (CAN)), naturally forming a distributed sensor system. An illustrative example is shown in Fig.~\ref{fig_lidar_configuration_1}, where 5 LiDARs are installed on a robotic ground vehicle moving in 6 DoF. 
	
	The use of multiple distributed LiDARs brings many new challenges in its localization and mapping: (1) extrinsic calibration. Since LiDARs are installed at different (and usually far apart) locations of the vehicle body, their relative pose is not perfectly rigid and may drift over time. This requires an online extrinsic calibration. This will be even more challenging  when two adjacent LiDARs have very small overlap; (2) high network bandwidth and computation requirements. A LiDAR is usually generating raw point data at a very fast rate. Sending all LiDAR data to a central computer could not only easily jam the vehicle network and the central computer, causing single point of failure, but also dramatically increases the computation power (meaning high power consumption, large noise, etc.). A potentially more robust way is to process each LiDAR's point data in its dedicated computer (e.g., electronic computing unit) and communicate the processed data (e.g., vehicle state, which is usually very small data) via the vehicle network. 
	
	In this paper, we present a decentralized multi-LiDAR calibration, localization and mapping system. The system is based on a decentralized formulation of EKF algorithm, which simultaneously runs on all LiDAR computers (or its allocated computing resources). All EKF copies perform the same procedures: maintaining an augmented state vector consisting the pose (and velocity) of the geometric center and the extrinsic parameters of all LiDARs, predicting from the most recent state update received from the rest EKF copies in the network, updating the state vector with new coming frames from its local LiDAR, and publishing the updated state vector to the network for other EKF copies to use.  
	
	In summary, our contributions are:
	\begin{itemize}
		\item We have proposed a calibration, localization and mapping system utilizing constant velocity model and EKF, which is capable of online estimating and updating LiDAR extrinsic w.r.t. geometric center.
		
		\item We present a decentralized multi-sensor calibration and fusion framework, which could be implemented in a distributed way and are potentially more robust to failures of central computers or individual sensors. 
		
		\item We have verified the convergence and accuracy of the proposed framework on actual systems and have achieved high precision localization and mapping results when compared with previous single LiDAR SLAM solution. 
	\end{itemize}
	
	\section{Related work}
	
	To date, multi-LiDAR sensors have been implemented in obstacle detection and tracking~\cite{sualeh2019,zeng2013}, computing occupancy map~\cite{huang2019} and natural phenomenon observation~\cite{newman2016,kopp2003}. All these setups rely on a central processing unit for computation and data exchange. Few research attention has been focused on the decentralization property of the multi-LiDAR system, however, which makes the above mentioned applications vulnerable to sensor message delay or loss. Furthermore, the unsupervised extrinsic calibration and sensor fusion of multi-LiDAR system remain to be discovered.
	
	
	Combining several sensors has led to the issue of multi-sensor data fusion. The simplest way is to use loosely coupled sensor fusion~\cite{lynen2013}, though computationally efficient, the decoupling of multi-sensor constraints may cause information loss. Tightly coupled sensor fusion model have also been discussed in~\cite{leutenegger2013} to improve the map accuracy. The joint optimization of entire sensor measurements and state vector is too time consuming, however, especially for high frequency sensor like LiDAR. Originated from statistics, EKF based sensor fusion~\cite{zhang2014loam} has become dominant in LiDAR SLAM due to their simple, linear and efficient recursive algorithm. In our approach, we implement EKF to maintain an augmented state vector to achieve a balance between productivity and precision.
	
	
	In addition to filtering, extrinsic calibration (recover rigid-body transformation between sensors) has been widely implemented to improve the SLAM precision. The majority of current LiDAR extrinsic calibration involve the following assumptions: known retroreflective targets or artificial environments~\cite{muhammad2010}. This requirement is hard to meet if users want to customize the mounting position that unsupervised calibration is preferred. Motion based approaches have been described in~\cite{taylor2016motion}, however, their results are easily affected by the cumulated drift from the motion. Appearance based approaches have been addressed in\cite{levinson2014unsupervised} that the optimal extrinsic is solved by maximizing overall point cloud quality. In contrast, our approach starts from a given initial value and iteratively utilizes EKF to refine extrinsic online. To the best our knowledge, our work is the first work that fuses data from multiple LiDARs in a decentralized framework, which can not only address the problem of localization and mapping, but can also online calibrate the extrinsic of 6-DoF. The results shown in Section. \ref{sect_res} demonstrates that our approach is of high-precision and effectiveness.
	
	
	\section{Overview}
	\begin{figure}[t]
		{
			\includegraphics[width=1.0\columnwidth]{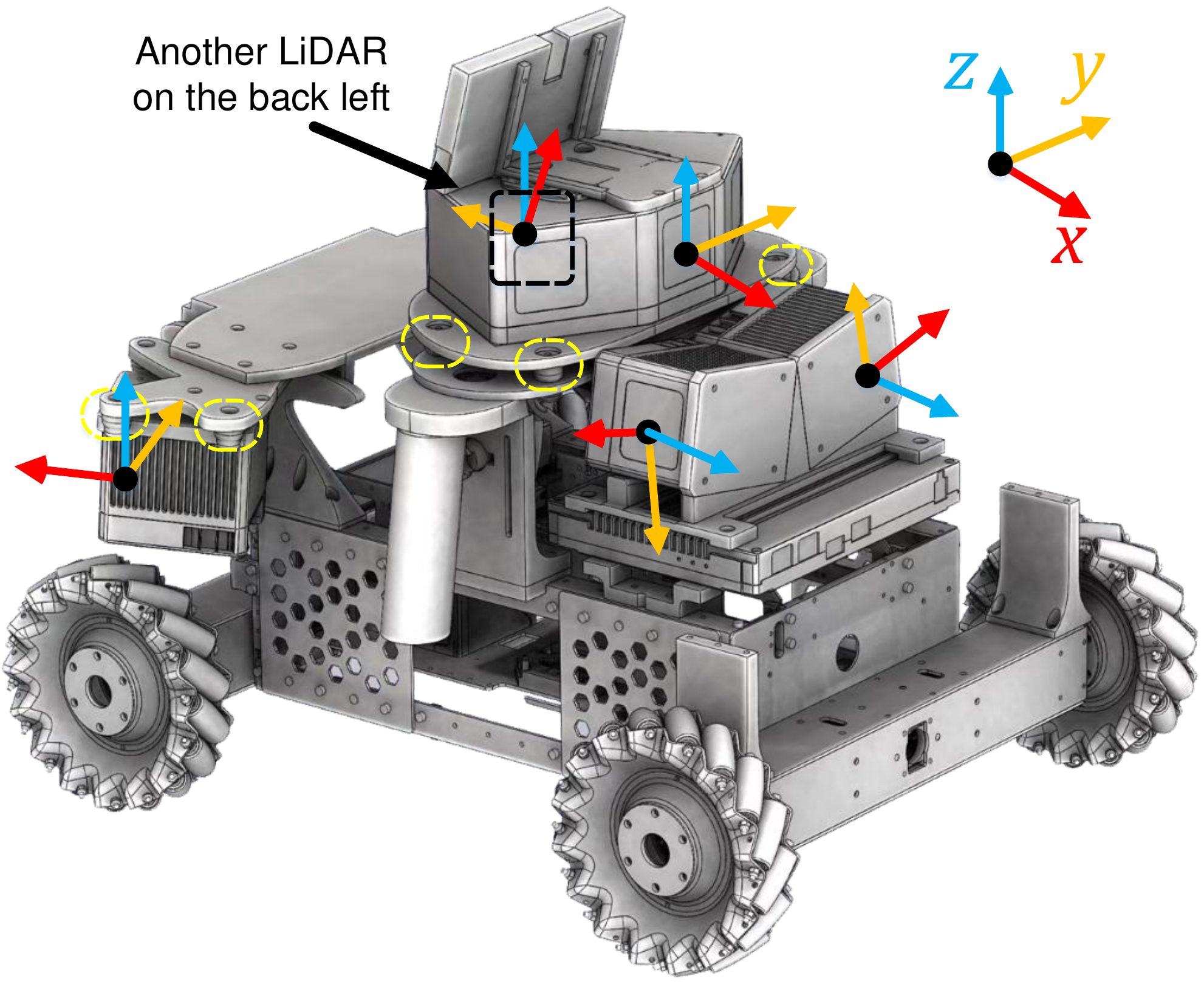}
			\caption{Our platform for data sampling, with 5 LiDARs installed on the car platform. To prevent the mechanical vibration caused by the rotating wheels on the rough ground, we add some damper ball between the connection of LiDAR and the platform (marked inside the yellow dashed circle).}
			\label{system_config}
			\vspace{-1.0cm}
		}
	\end{figure}
\begin{figure}[t]
		{	
			\centering
			{\includegraphics[width=0.8\columnwidth]{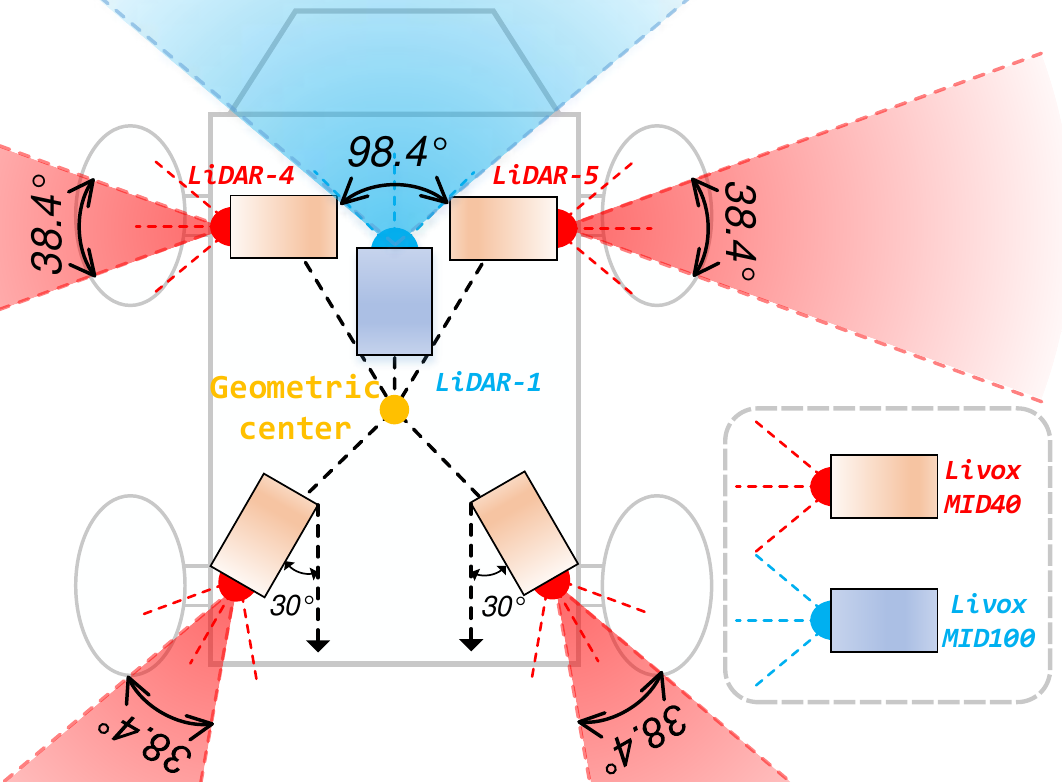}
				\caption{The configurations of our LiDAR installation.}
				\label{fig_lidar_configuration_2}
			}
			\vspace{-0.0cm}
		}
	\end{figure}

	The configuration of our system is shown in Fig. \ref{system_config}, we have five different LiDARs installed on the car platform, with their front face looking ``front'', ``left'', ``right'', ``back-left'' and ``back-right''. LiDAR-$1$ is Livox-MID100\footnote{\url{https://www.livoxtech.com/mid-40-and-mid-100/specs}}, with $98.4^\circ$ of horizontal and $38.4^\circ$ of vertical FoV. Other LiDARs are all Livox-MID40, with $38.4^\circ$ circular FoV. Due to the limited FoV, there is no overlapping areas between any two LiDARs (see Fig. \ref{fig_lidar_configuration_2}).
	
	To prevent the vibrations caused by rotating the \textit{mecanum wheels}\footnote{\url{https://en.wikipedia.org/wiki/Mecanum\_wheel}} on rough ground, which could cause high-frequency motion blur effect on the LiDAR point cloud. We add the rubber damping-ball at the connection of LiDAR and car platform, which could effectively compensate the vibration. In addition, most of our mechanical parts are 3D printed with the $PLA$ material, which can be easily distorted by the applied force. By this, we do not treat our LiDAR group as a rigid system.
	
	Our platform is of low-cost, with all of our mechanical parts being 3D printed, whose total price is about 4k USD (details are shown in Table. \ref{table:cost_list}). With the algorithm proposed in our following sections, we can achieve a precision around $0.2\%$. For more details of our platform, we strongly recommend the readers visiting the project on our GitHub.\footnote{\url{https://github.com/hku-mars/lidar_car_platfrom}}
	
	\begin{table}
		\setlength{\extrarowheight}{.1em}
		\setlength\extrarowheight{0.01pt}
		\begin{center}
			\begin{tabular}[H]{|c|c|c|c|c|}
				\hline
				& Livox & Livox & 3D printing  &  Total\\
				& MID100 & MID40 & \& others   &       \\
				\hline
				Price& \$ 1499 & \$ $ 599\times 4 $ & $\leq \$ 100$ & $\leq \$3995 $ \\
				\hline
			\end{tabular}	
		\end{center}
		\caption{List of the materials (not including the robot platform) in our multi-LiDAR system, with the total price about 4k USD, including 5 LiDARs and the fees of 3D printing. }
		\label{table:cost_list}
		\vspace{-1.5cm}
	\end{table}
	
	\section{Vehicle model}
	
	\subsection{Notation}
	In our work, we use the $4\times 4$ matrix $\textbf{T}$ to denote the pose and rigid transformation in 6 degrees of freedom (DoF):
	$$
	\textbf{T} =\begin{bmatrix}
	\textbf{R} & \textbf{t} \\
	\textbf{0} & 1
	\end{bmatrix} \in SE(3)
	$$
	
	Since the $3\times 3$ rotation matrix $\textbf{R}$ is of 3 DoF, we use a rotation vector $ \textbf{r} \in \mathbb{R}^{3} $ as a minimal parameterization, which represents the rotation in the form of angle$-$axis.
	\begin{align}
	\textbf{R}  = e^{\widehat{\textbf{r}}} &
	\in SO(3)
	\end{align}
	where $\widehat{\cdot}$ transforms a vector into a skew-symmetric matrix. The conversions between  $\textbf{R}$ and  $\textbf{r}$ are denoted as $\textbf{R}\{\cdot\}$ and $\textbf{r}\{\cdot\}$ for convenience.
	\begin{align}
	&\textbf{R} = \exp{\left(\textbf{r}\right)}  = \textbf{R} \left( \textbf{r} \right)  \\
	&\textbf{r} = \log{\left(\textbf{R}\right)}  = \textbf{r} \left( \textbf{R}  \right)
	\end{align}
	
	Using the minimal parameterization of $\mathbf{R}$, the full pose $\mathbf{T}$ can also be parameterized minimally by $\mathbf{x} = \left[\textbf{r}, \textbf{t}\right]^T \in \mathbb{R}^6$, and the conversions between these two can be denoted as:
	\begin{align}
	&\textbf{T} = \mathbf{T}{\left(\textbf{x}\right)}  \\
	&\textbf{x} = \mathbf{x}{\left(\textbf{T}\right)}
	\end{align}
	Sometimes, we also use the notation  $\mathbf{T} = \left( \textbf{r}, \textbf{t}\right)$ to represent the minimal parameterization of $\mathbf{T}$. 
	
	\subsection{Constant velocity model}\label{sect_constant_vel_model}
	
	Viewing the robot as a rigid body, its pose can be represented by a reference frame (e.g., at the geometric center in Fig. \ref{fig_lidar_configuration_2}). Furthermore, we use a constant velocity model as in \cite{davison2003real} to model the 6 DoF motion of the robot.  Denote $\textbf{r}^c_k$ the robot attitude , $ \textbf{t}^c_k$ is the translation, $\boldsymbol{\omega}^c_{k}$ the angular velocity, and $\textbf{v}^c_{k}$ the linear velocity, all at time $k$, then a constant velocity model yields a state equation as below:
	\begin{align}
	\textbf{r}^c_{k+1} &= \textbf{r} \left( \mathbf{R} \left( \textbf{r}^c_k  \right)  \exp{\left(\widehat{\boldsymbol{\omega}}^c_{k}\cdot\Delta t \right)} \right) \\
	\textbf{t}^c_{k+1} &= \textbf{t}^c_{k} + \textbf{v}^c_{k} \cdot \Delta t \\
	\boldsymbol{\omega}^c_{k+1} &= \boldsymbol{\omega}^c_{k} + \boldsymbol{ \epsilon }_\omega \\
	\textbf{v}^c_{k+1} &= \textbf{v}^c_{k} + \boldsymbol{ \epsilon }_\textbf{v} 
	\end{align}
	where $\Delta t$ is the time difference from the last update at $t_k$ and the current update at $t_{k+1}$ (i.e., $\Delta t = t_{k+1} - t_{k}$), $\boldsymbol{ \epsilon }_\omega$ and $\boldsymbol{ \epsilon }_\textbf{v}$ are the force and torque impulse applied to the robot. They are usually modeled as zero-mean Gaussian noise:
	$$
	\left[ \boldsymbol{ \epsilon }_\omega, \boldsymbol{ \epsilon }_\textbf{v}\right]^T \sim \mathcal{N}(\mathbf{0}, \boldsymbol{\Sigma}_{\mathbf{w}})
	$$
	
	The above state model can be rewritten as a more compact form as below:
	\begin{align}
	\textbf{x}^c_{k+1} = \mathbf{f} \left( \textbf{x}^c_{k}, \textbf{w}; \Delta t \right) \in \mathbb{R}^{12}
	\end{align}
	where $\textbf{x}^c_{k} = \left[\textbf{r}^c_{k}, \textbf{t}^c_{k}, \boldsymbol{\omega}^c_{k}, \textbf{v}^c_{k} \right]^T $ and $\textbf{w} = \left[ \boldsymbol{ \epsilon }_\omega, \boldsymbol{ \epsilon }_v \right]^T$.
	
	\subsection{Extrinsic model}\label{sect_extrinsic}
	Assuming there are $N$ LiDARs and $\textbf{T}^{ei} = \left( \textbf{r}^{ei}, \textbf{t}^{ei} \right)$ denotes the extrinsic of $i$-th LiDAR frame w.r.t the reference frame, we have the pose $\textbf{T}^i_{k}$ of $i$-th LiDAR at time $k$ as:
	\begin{align}
	\textbf{T}^i_{k} &=  \textbf{T}_{k}^c \textbf{T}_{k}^{ei}  \nonumber \\
	&=\begin{bmatrix} 
	\textbf{R}^c_{k} \textbf{R}^{ei}   & \textbf{R}_k^c \textbf{t}^{ei} +  \textbf{t}^c_{k}\\
	\textbf{0} & 1
	\end{bmatrix} \label{extrinsic_model}
	\end{align}
	where $\textbf{T}^{c}_k = \left( \textbf{r}^{c}_k, \textbf{t}^{c}_k \right)$ is the pose at time $t_k$. 

	\subsection{Full state model}\label{Sect_full_state}
	
	Denote $\textbf{x}^{ei} = \left[ \textbf{r}^{ei}, \textbf{t}^{ei} \right]^T \in \mathbb{R}^{6}$ as the state associated the $i$-th LiDAR extrinsic parameters, then the full state is 
	\begin{align}
	\label{e:full_state_def}
	\textbf{x} =
	\begin{bmatrix}
	\textbf{x}^c & \textbf{x}^{e1} & \textbf{x}^{e2} & \cdots & \textbf{x}^{eN}
	\end{bmatrix}^T \in \mathbb{R}^{12 + 6N}
	\end{align}
	\noindent
	and the state model is
	\begin{align}
	\textbf{x}^c_{k+1} &= \mathbf{f} \left( \textbf{x}^c_{k}, \textbf{w}; \Delta t \right) \\
	\textbf{x}^{e1}_{k+1} &=\textbf{x}^{e1}_{k} \\
	\textbf{x}^{e2}_{k+1} &=\textbf{x}^{e2}_{k} \\
	\vdots \\
	\textbf{x}^{e2}_{k+1} &= \textbf{x}^{eN}_{k}
	\end{align}
	
	Notice that state vector $\textbf{x}$ in (\ref{e:full_state_def}) retains all information for determining the robot state in the future, therefore being a valid state. For example, the $i$-th LiDAR pose can be determined from $\textbf{x}$ as follows:
	\begin{align}
	\textbf{T}^i_{k+1} &= \textbf{T} \left( \mathbf{x}_{k}^c \right) \textbf{T} \left(  \mathbf{x}_{k}^{ei} \right) 
	\end{align}

	\begin{figure}[t]
		\includegraphics[width=1.00\columnwidth]{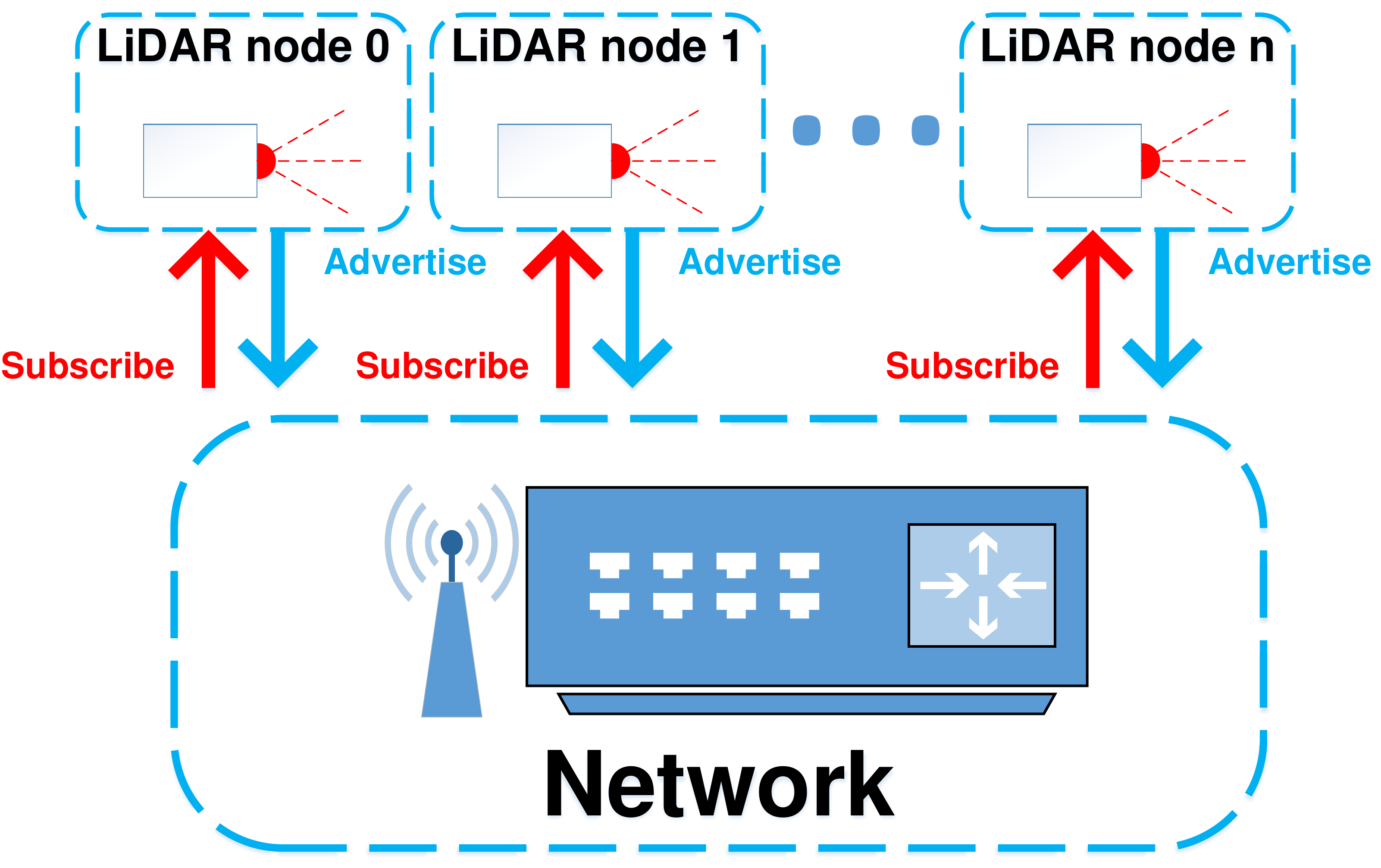}
		\caption{The framework of our decentralized system, each LiDAR refreshes the newest state $\textbf{x}$ by subscribing the message from the network. Once the point cloud registration with latest coming data is complete, it will advertise the selected feature points and the updated state to the network.}
		\label{fig_decenterlize_framework}
		\vspace{-1.0cm}
	\end{figure}	
	
	\subsection{Measurement model}\label{sect_model_measure}
	Our decentralized EKF runs a LiDAR odometry and mapping (LOAM) algorithm \cite{lin2019} for each LiDAR on its dedicated computing devices, usually an onboard computer with modulate computing performance or a virtual computation resource allocated from a high performance server. Taking the $i$-th LiDAR as an example, the LOAM solves the $i$-th LiDAR pose at time $t_{k+1}$ (i.e., $\textbf{T}^i_{k+1}$) by minimizing the distance of edge features $\textbf{r}_{p2e}$ and plane features $\textbf{r}_{p2p}$ between the current scan and a local map
	\begin{align}
	\label{e:opt_original}
	\mathop{\min}_{\textbf{T}^i_{k+1} \in SE(3)}  \left( \sum{\textbf{r}_{p2e}} + \sum{\textbf{r}_{p2p}} \right) 
	\end{align}
	
	To accelerate the optimization process, the predicted $i$-th LiDAR pose $\bar{\mathbf{T}}^i_{k+1}$ from \textit{Section. \ref{sect_ekf_predict}} is usually used as the initial estimate, and the error pose from which is solved. i.e.,
	\begin{align}
	\label{e:meas_err_model}
	\textbf{T}^i_{k+1} &= \bar{\textbf{T}}^i_{k+1} \textbf{T} (\delta \mathbf{x}^i_{k+1})
	\end{align}
	
	Substituting this into (\ref{e:opt_original}) leads to 
	\begin{align}
	\label{e:opt_delta}
	\delta \widehat{\mathbf{x}}^i_{k+1} = \mathop{\arg \min}_{ \delta \mathbf{x}^i_{k+1} \in \mathbb{R}^ 6 }  \left( \sum{\textbf{r}_{p2e}} + \sum{\textbf{r}_{p2p}} \right) 
	\end{align}
	Assume the Hessian matrix of (\ref{e:opt_delta}) at convergence is $\hat{\boldsymbol{ \Sigma}}_{\delta}^{-1}$, then $\hat{\boldsymbol{ \Sigma}}_{\delta}$ is the covariance matrix associated to the measurement $\delta \widehat{\mathbf{x}}^i_{k+1}$. That is to say, the measurement model is
	\begin{align}
	\label{e:icp_meas_model}
	\delta \widehat{\mathbf{x}}^i_{k+1} & = \delta {\mathbf{x}}^i_{k+1} + \mathbf{v}
	\end{align}
	where $\mathbf{v} \sim \mathcal{N}(\mathbf{0}, \hat{\boldsymbol{ \Sigma}}_{\delta}) $ and $\delta {\mathbf{x}}^i_{k+1}$ is solved from (\ref{e:meas_err_model})
	\begin{align}
	\delta {\mathbf{x}}^i_{k+1} &= \mathbf{x}\left( \left( \bar{\textbf{T}}^i_{k+1} \right)^{-1} \textbf{T} \left( \mathbf{x}_{k}^c \right) \textbf{T} \left(  \mathbf{x}_{k}^{ei} \right)  \right)
	\end{align}
	notice that $\delta {\bar{\mathbf{x}}}^i_{k+1} = \mathbf{0}$.

	\section{Decentralized Extended Kalman Filter}\label{Sect_const_velo_ekf}
	In this section, we introduce our decentralized EKF algorithm. Unlike existing EKF algorithm which often runs a single instance on a central computer, our system runs multiple EKF instances in parallel, one per LiDAR. Individual instance usually runs on the respective LiDAR dedicated computing resources and is responsible for processing that LiDAR data. As shown in Fig. \ref{fig_decenterlize_framework}, each EKF reads the full state vector $\textbf{x} =
	\begin{bmatrix}
	\textbf{x}^c & \textbf{x}^{e1} & \textbf{x}^{e2} & \cdots & \textbf{x}^{eN}
	\end{bmatrix}^T$ from the network, updates it by registering the respective LiDAR data, and publishes the updated state to the network for other EKF instances to use. In the following, we explain in detail how the full state is updated for each LiDAR (e.g., $i$-th LiDAR). 
	
	\begin{figure}[t]
		\includegraphics[width=1.00\columnwidth]{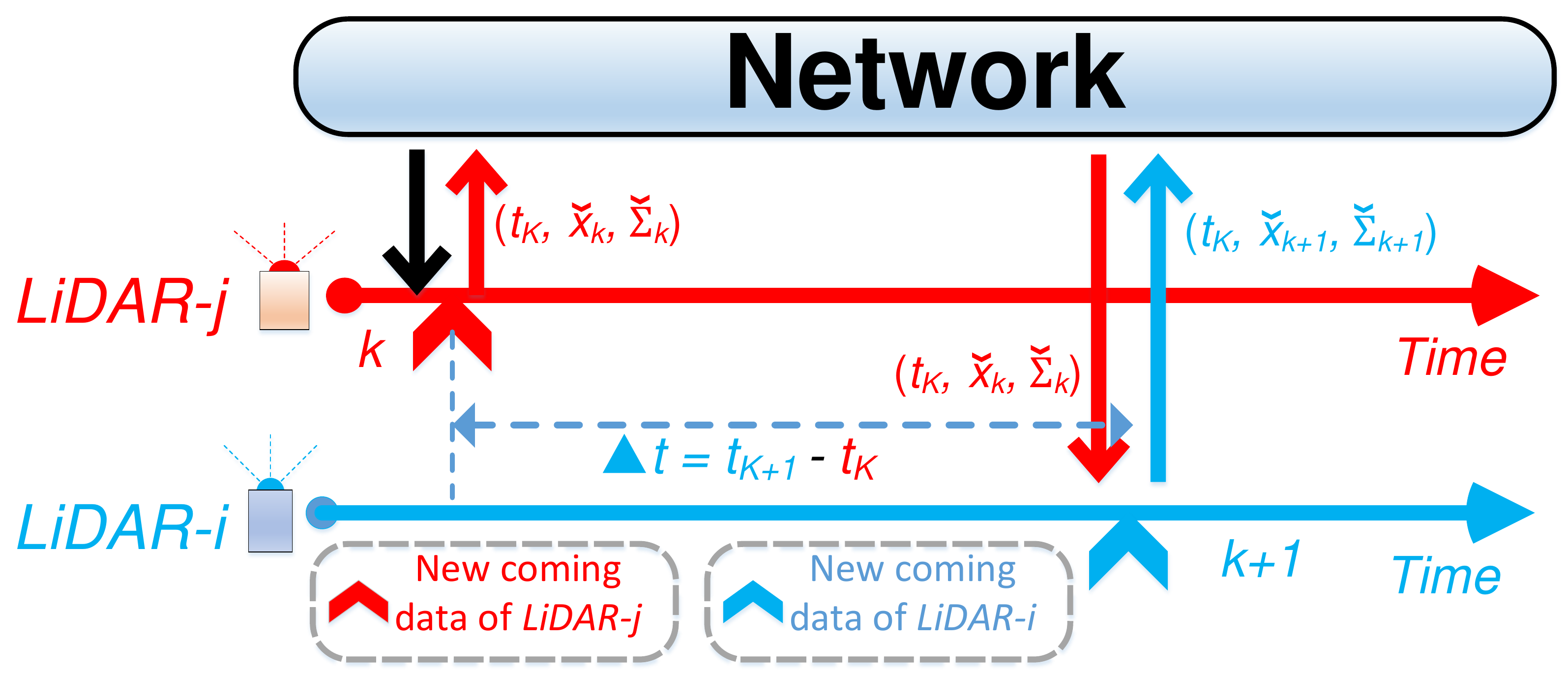}
		\caption{One step update of our decentralized EKF algorithm: once receiving a point cloud scan at time $t_{k+1}$, the $i$-th LiDAR retrieves the newest state update  $(\check{\textbf{x}}_k, \check{\boldsymbol{ \Sigma}}_k )$, which was received from the network at its local time $t_{k}$. Then it uses its scan to update the state, and advertise the updated state $(\check{\textbf{x}}_k, \check{\boldsymbol{ \Sigma}}_{k+1} )$ to the network.}
		\vspace{-0.8cm}
		\label{fig_decenterlize_example}
	\end{figure}
	
	\subsection{State prediction}\label{sect_ekf_predict}
	As shown in Fig. \ref{fig_decenterlize_example}, assume the $i$-th LiDAR obtains a scan at time $t_{k+1}$. Moreover, assume the most recent state update $\check{\textbf{x}}_k$ (and associated covariance $\check{\boldsymbol{ \Sigma}}_k$) was published by LiDAR $j$ ($j$ could be equal to $i$). The $\check{\textbf{x}}_k$ and $\check{\boldsymbol{ \Sigma}}_k$ were received by the LiDAR $i$ at its local time $t_k$ ($t_k < t_{k+1}$).
	
	Refer to the Section \ref{Sect_full_state}, we have: 
	\begin{align}
	\check{\textbf{x}}_k =
	\begin{bmatrix}
	\check{\textbf{x}}^c_k & \check{\textbf{x}}^{e1}_k & \check{\textbf{x}}^{e2}_k & \cdots & \check{\textbf{x}}^{eN}_k
	\end{bmatrix}^T
	\end{align}
	
	Then, follow the standard EKF prediction, we have the predicted full state vector:
	\begin{align}
	\bar{\textbf{x}}_{k+1} =
	\begin{bmatrix}
	\bar{\textbf{x}}^c_{k+1} & \bar{\textbf{x}}^{e1}_{k+1} & \bar{\textbf{x}}^{e2}_{k+1} & \cdots & \bar{\textbf{x}}^{eN}_{k+1}
	\end{bmatrix}^T \label{eq_dec_ekf_x_k1}
	\end{align}
	computed as below:
	\begin{align}
	\bar{\textbf{x}}^c_{k+1} &= \mathbf{f} \left( \check{\textbf{x}}^c_{k}, \textbf{0}; \Delta t \right) \label{x_x_e2_xc} \\
	\bar{\textbf{x}}^{e1}_{k+1} &=\check{\textbf{x}}^{e1}_{k} \\
	\vdots \nonumber \\
	\bar{\textbf{x}}^{eN}_{k+1} &= \check{\textbf{x}}^{eN}_{k} \label{x_x_e2_x_en}
	\end{align}
	where $\Delta t = t_{k+1}- t_{k}$. The covariance matrix associated to the state prediction $\bar{\textbf{x}}_{k+1}$ is
	\begin{align}
	\bar{\boldsymbol{ \Sigma}}_{k+1} = \textbf{F} \check{\boldsymbol{ \Sigma}}_k \textbf{F}^T + \textbf{G} \boldsymbol{\Sigma}_{\mathbf{w}} \textbf{G}^T \label{eq_sigma_c_k1}
	\end{align}
	where $\textbf{F} \in \mathbb{R}^{(12+6N)\times (12+6N)}$ and $\textbf{G} \in \mathbb{R}^{(12+6N)\times 6}$
	\begin{align}
	\textbf{F} = 
	\begin{bmatrix}
	\dfrac{\partial \mathbf{f} \left( {\textbf{x}}^c_{k}, \textbf{0}; \Delta t \right)}{\partial {\textbf{x}}^c_{k}} & \textbf{0} \\
	\textbf{0} & \textbf{I}_{6N} \\
	\end{bmatrix} , \ \ \textbf{G} = \dfrac{\partial \mathbf{f} \left( \check{\textbf{x}}^c_{k}, \textbf{w}; \Delta t \right)}{\partial {\textbf{w}}} 
	\end{align}
	
	Then we can predict the pose of $i$-th LiDAR pose $\bar{\textbf{T}}^i_{k+1}$ at time $t_{k+1}$:
	\begin{align}
	\bar{\textbf{T}}^i_{k+1} = \textbf{T} \left( \bar{\mathbf{x}}_{k+1}^c \right) \textbf{T} \left(  \bar{\mathbf{x}}_{k+1}^{ei} \right) \label{eq_dec_ekf_k1}
	\end{align}
	which is used as the initial estimate of $\mathbf{T}^i_{k+1}$ used in the LOAM as explained in \textit{Section. \ref{sect_model_measure}}.
	\subsection{Measurement update}
	
	A problem with the state equation  (\ref{x_x_e2_xc}$\sim$\ref{x_x_e2_x_en}) is that it involves $N$ extrinsic parameters. With measurements $\delta {\widehat{\mathbf{x}}}^i_{k+1}$ from the point registration in (\ref{e:opt_delta}), the system is not observable (nor detectable), causing the EKF to diverge. This problem is usually resolved by fixing the reference frame at any one of the $N$ LiDARs, removing the extrinsic estimation of that LiDAR. However, when the reference LiDAR fails, the rests have to agree on another reference LiDAR, which is usually a  complicated process. 
	
	To avoid this, we choose the reference frame at the center of all LiDARs. i.e.,
	\begin{align}
	\mathbf{t}^c_k &= \frac{1}{N} \sum_{i=1}^{N} \mathbf{t}^i_k ; \ \forall k = 0, 1, 2, \cdots
	\end{align}
	
	Substituting in (\ref{extrinsic_model}) leads to
	\begin{align}
	\sum_{i=1}^{N} \textbf{t}^{ei}  = \mathbf{0}
	\end{align}
	
	Besides the location, the attitude of the reference frame $\mathbf{R}^c_k$ are defined such that
	\begin{align}
	\sum_{i=1}^{N} \mathbf{r}\left( \left( \mathbf{R}^c_k \right)^T \mathbf{R}^i_k \right) = \mathbf{0}
	\end{align}
	which leads to
	\begin{align}
	\sum_{i=1}^{N} \textbf{r}^{ei}  = \mathbf{0}
	\end{align}
	
	As a result, in addition to the measurement $\delta {\widehat{\mathbf{x}}}^i_{k+1}$ in (\ref{e:icp_meas_model}), two new measurements of $\mathbf{0}_{3 \times 1}$ should be added. The total measurement vector is
	\begin{align}
	\mathbf{y}_m = \left[\delta {\widehat{\mathbf{x}}}^i_{k+1}, \mathbf{0}, \mathbf{0} \right]^T
	\end{align}
	and the respective output functions are
	\begin{align}
	\label{e:extrinsic_output}
	\textbf{r}_{re} =  \sum_{i=1}^{N} \textbf{r}^{ei}, \hspace{0.2cm} \textbf{t}_{re} = \sum_{i=1}^{N} \textbf{t}^{ei} 
	\end{align}
	
	Then we have the residual vector $\textbf{z}$ and the associated covariance $\boldsymbol{ \Sigma}_z$ are
	\begin{align}
	\textbf{z} &= \begin{bmatrix}
	\delta\widehat{\mathbf{x}}^i_{k+1} & -\bar{\textbf{r}}_{re} & -\bar{\textbf{t}}_{re}
	\end{bmatrix}\in \mathbb{R}^{12} \\
	\boldsymbol{ \Sigma}_z &=
	\begin{bmatrix}
	\hat{\boldsymbol{ \Sigma}}_{\delta} & \textbf{0} \\
	\textbf{0} & s\textbf{I}
	\end{bmatrix}\in \mathbb{R}^{12\times 12}
	\end{align}
	where $s$ is a small value to prevent the EKF from being singular (set as $1$ in our work), $\bar{\textbf{r}}_{re}$ and $ \bar{\textbf{t}}_{re}$ are the sum of predicted extrinsic rotation and translation in (\ref{e:extrinsic_output}), respectively. 
	
	As a result, the Kalman gain is
	\begin{align}
	\textbf{K} &= \bar{\boldsymbol{ \Sigma}}_{k+1}\textbf{H}^T\left( \textbf{H}\bar{\boldsymbol{ \Sigma}}_{k+1}\textbf{H}^T + \boldsymbol{ \Sigma}_z \right)^{-1} \in \mathbb{R}^{\left(12+6N\right)\times 12} \label{eq_c_kalman_gain}
	\end{align}
	\noindent
	with $\textbf{H} \in \mathbb{R}^{12\times \left(12+6N\right)} $ being
	\begin{align}
	\textbf{H} =
	\begin{bmatrix}
	\dfrac{ \partial ( \delta {\mathbf{x}}^i_{k+1} )}{ \partial {\textbf{x}}^c_{k}  } & \textbf{0} & \dots & \dfrac{ \partial ( \delta {\mathbf{x}}^i_{k+1})}{ \partial {\textbf{x}}^{ei}  } & \dots & \textbf{0} \\ 
	\textbf{0} & \dfrac{\partial \textbf{r}_{re} }{ \partial {\textbf{x}}^{e1}} & \dots & \dfrac{\partial \textbf{r}_{re} }{ \partial {\textbf{x}}^{ei}} &\dots & \dfrac{\partial \textbf{r}_{re} }{ \partial {\textbf{x}}^{eN}} \vspace{0.1cm}\\
	\textbf{0} & \dfrac{\partial \textbf{t}_{re} }{ \partial {\textbf{x}}^{e1}} & \dots & \dfrac{\partial \textbf{t}_{re} }{ \partial {\textbf{x}}^{ei}}  &\dots &  \dfrac{\partial \textbf{t}_{re} }{ \partial {\textbf{x}}^{eN}}
	\end{bmatrix}
	\end{align}
	
	Finally, we have the measurement update as follows: 
	\begin{align}
	\check{\textbf{x}}_{k+1} &= \bar{\textbf{x}}_{k+1} + \textbf{K}\textbf{z} \label{eq_x_c_star_k1}\\
	\check{\boldsymbol{ \Sigma}}_{k+1} &= \left( \textbf{I} - \textbf{K}\textbf{H} \right)\bar{\boldsymbol{ \Sigma}}_{k+1} \label{eq_sig_c_star_k1}
	\end{align}
	The updated full state is then advertised to the network for other EKF instances to use. 
	
	\subsection{Algorithm of decentralized calibration, localization, and mapping with multiple LiDARs}
	To sum up, we conclude the previous EKF formulation as the algorithm shown below:
	\begin{algorithm}
		\caption{Decentralized calibration, localization and mapping on the $i$-th LiDARs}
		\label{alg_it_pose}
		\renewcommand{\thealgocf}{}
		\renewcommand{\theAlgoLine}{}  
		\SetKwInOut{Input}{Input}
		\SetKwInOut{Output}{Output }
		\SetKwInOut{Begin}{Begin}
		\SetKwInOut{Start}{Start}
		\SetKwInOut{Prediction}{Prediction}
		\SetKwInOut{Mes}{Update}
		\SetKwInOut{Return}{Return}
		\Input{ $\check{\textbf{x}}_{k}$, $\check{\boldsymbol{ \Sigma}}_{k}$ received from the network at time $t_k$;  Current point cloud of $i$-th LiDAR received at time $t_{k+1}$.}
		\Output{Advertise the updated state $\check{\textbf{x}}_{k+1}$ and its associated covariance matrix $\check{\boldsymbol{ \Sigma}}_{k+1}$ to the network. }
		\Prediction
		{\\
			Get $ \bar{\textbf{x}}_{k+1} $ from  (\ref{eq_dec_ekf_x_k1}).\\
			Get $ \bar{\boldsymbol{ \Sigma}}_{k+1} $ from (\ref{eq_sigma_c_k1}).\\
			Compute $\bar{\textbf{T}}^i_{k+1}$ from (\ref{eq_dec_ekf_k1}).
		}
		\Mes
		{\\
			Solve $\delta\widehat{\mathbf{x}}^i_{k+1}$ and $\hat{\boldsymbol{ \Sigma}}_{\delta}$ from (\ref{e:opt_delta}). \\
			Get the Kalman gain $\textbf{K}$ from (\ref{eq_c_kalman_gain}). \\
			Update $\check{\textbf{x}}_{k+1}$ from (\ref{eq_x_c_star_k1}).\\
			Update $\check{\boldsymbol{ \Sigma}}_{k+1}$ from (\ref{eq_sig_c_star_k1}).
		}
		\Return
		{
			$\check{\textbf{x}}_{k+1}$, $\check{\boldsymbol{ \Sigma}}_{k+1}$
		}
	\end{algorithm}
	
	\subsection{Initialization}
	\subsubsection{Hand-eye calibration}
	To provide the well initialized extrinsic, we implement the hand-eye calibration algorithm introduced in \cite{jiao2019automatic}. However, due the damper ball and the 3D-printed modules, the rigid connection is not guaranteed. We do not assume the extrinsic result are well calibrated every time, however, we believe it is suitable to serve as the initial guess at the beginning of EKF and map alignment.
	
	\subsubsection{Map alignment}
	Map alignment can not only provide the initial estimation of extrinsic among LiDARs, but also can align the different coordinates of LiDARs, making the odomety of each LiDARs to the same reference frame.
	
	Since there is no overlapping area among any two LiDARs, we can not directly find out the relative transformation between any two LiDARs. In our work, in the stage of initialization, each LiDAR node performs LOAM at their own frame coordinates, meanwhile, subscribes to the point cloud data published by others. Since the platform is moving, the mappings of LiDARs will have overlaps with others. Once the overlapping area is sufficient, the ICP algorithm is performed and we can align both the map and coordinate frames of each LiDAR.
	
	\section{Experiments}\label{sect_experiments}
\begin{figure}
		\includegraphics[width=1.00\columnwidth]{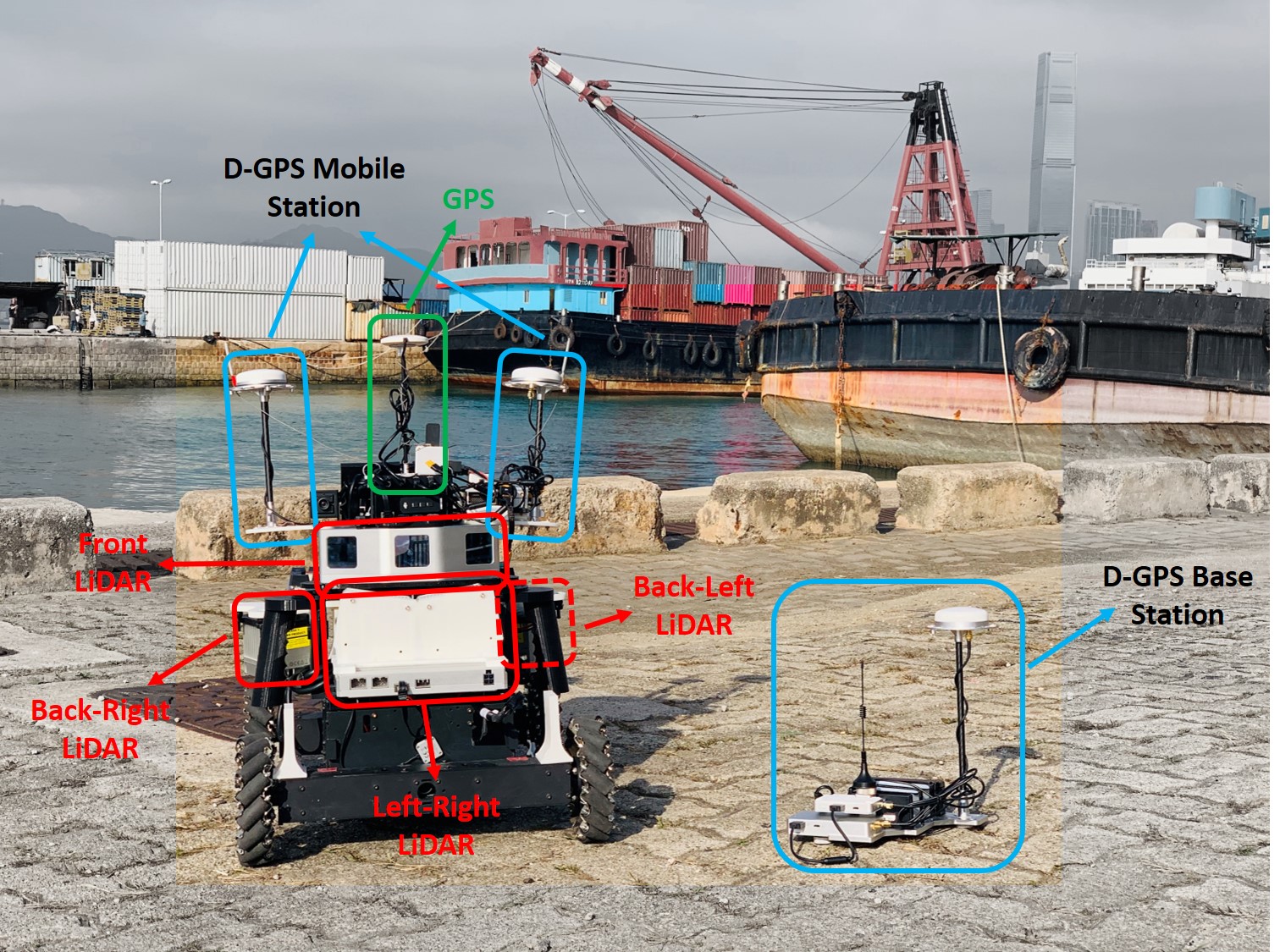}
		\caption{Our remotely operated vehicle platform consisting 5 LiDARs, onboard mini-computer, D-GPS mobile station and monocular camera (for pilot preview only).}
		\label{fig_lidar_configuration}
		\vspace{-0.5cm}
	\end{figure}

	Our custom-built robotic platform is shown in the Fig. \ref{fig_lidar_configuration}, with a Differential Global Positioning System (D-GPS) mobile station mounted on the top, which is used to provide a high precision odometry reference to evaluate our algorithm. We implement our decentralized framework on a high-performance PC, which is embedded with Intel i7-9700K processor and 32GB RAM. Similar to a real distributed system, in our implementation each LiDAR EKF algorithm runs in an individual ROS node by publishing and subscribing messages from each other. 
	
	We ran our vehicle platform at a harbour area with relative constant speed, good GPS signal and some moving pedestrians. The satellite image of our test ground is shown in Fig.~\ref{fig_maps}.B. Two trajectories, \textit{Scene-1} and \textit{Scene-2}, were recorded taking about $400s$ and $320s$ respectively. \textit{Scene-1} is a one-way trajectory while in \textit{Scene-2}, we chose to walk in relative straight lines and returned to an end point close to where we began, as shown in Fig.~\ref{fig_comp_trajecties}.
	
	\section{Results}\label{sect_res}
	
	The 5 EKF copies maintain the same state vector and update it at different time (once receive the respective LiDAR data). In the following results, we collect each state estimate across all 5 EKFs and analyze its convergence over time as well as accuracy if ground truth is available (e.g., position).
	
	\subsection{Result of online extrinsic calibration}
\begin{figure}
		\includegraphics[width=1.0\columnwidth]{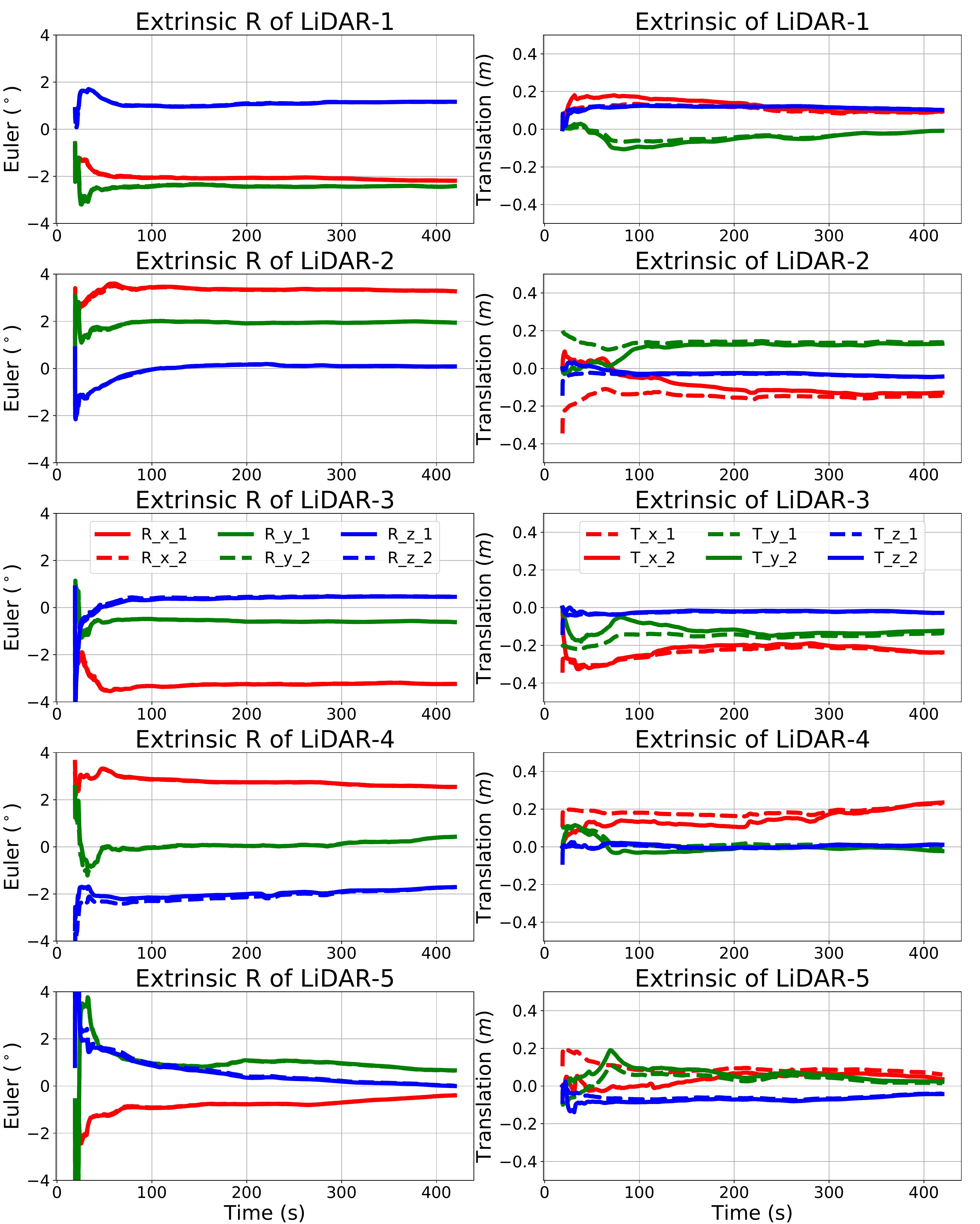}
		\caption{The updates of extrinsic parameters (left: rotation, right: translation) among 5 LiDARs with respect to the geometric center. The solid line starts with the initial guess calculated from map-alignment. The dashed line starts with zero initial values.}
		\label{fig_extrinsic}
		\vspace{-2.0cm}
	\end{figure}

	The extrinsic estimation of all 5 LiDARs with respect to the geometric center are plotted in Fig. \ref{fig_extrinsic}. The left and right columns depict the extrinsic of rotation in Euler angle and the translation in meter, respectively. We test our algorithm with two sets of initial values: the first one is obtained from the result of map alignment (solid line) and the second one is directly set to $0$ (dashed line).	As shown in Fig. \ref{fig_extrinsic}, both the extrinsic of rotation and translation converge to a stable value quickly, which demonstrates the ability of our algorithm to calibrate the extrinsic, even with inaccurate initial values. The converged extrinsic values also agrees with visual inspection of the location of each LiDAR. 
	
	\subsection{Result of state estimation}
	
	Besides the extrinsic parameters, we further present the state estimation of the vehicle geometric center. In \textit{Scene-1}, the pose and velocity estimation of the geometry are shown in Fig. \ref{fig_state_esimation_scene_1}, we can see that the estimation of velocity can reflect the change of poses very well. Taking the position of $x$-axis and its corresponding linear velocity for example, in the time interval $[18.9s, 300s] $, the value of \textit{Pos\_x} increases from $16m$ to about $400 m$, with constant velocity around $1.36 m/s$, which matches with the estimated velocity as shown in Fig. \ref{fig_state_esimation_scene_1}. The results of \textit{Scene-2} are similar and not presented here due to space limit. 
	
	\subsection{Evaluation of localization accuracy}
	
\begin{figure}[h]
		{	
			\vspace{-0.5cm}
			\includegraphics[width=0.495\columnwidth, height=4.5cm ]{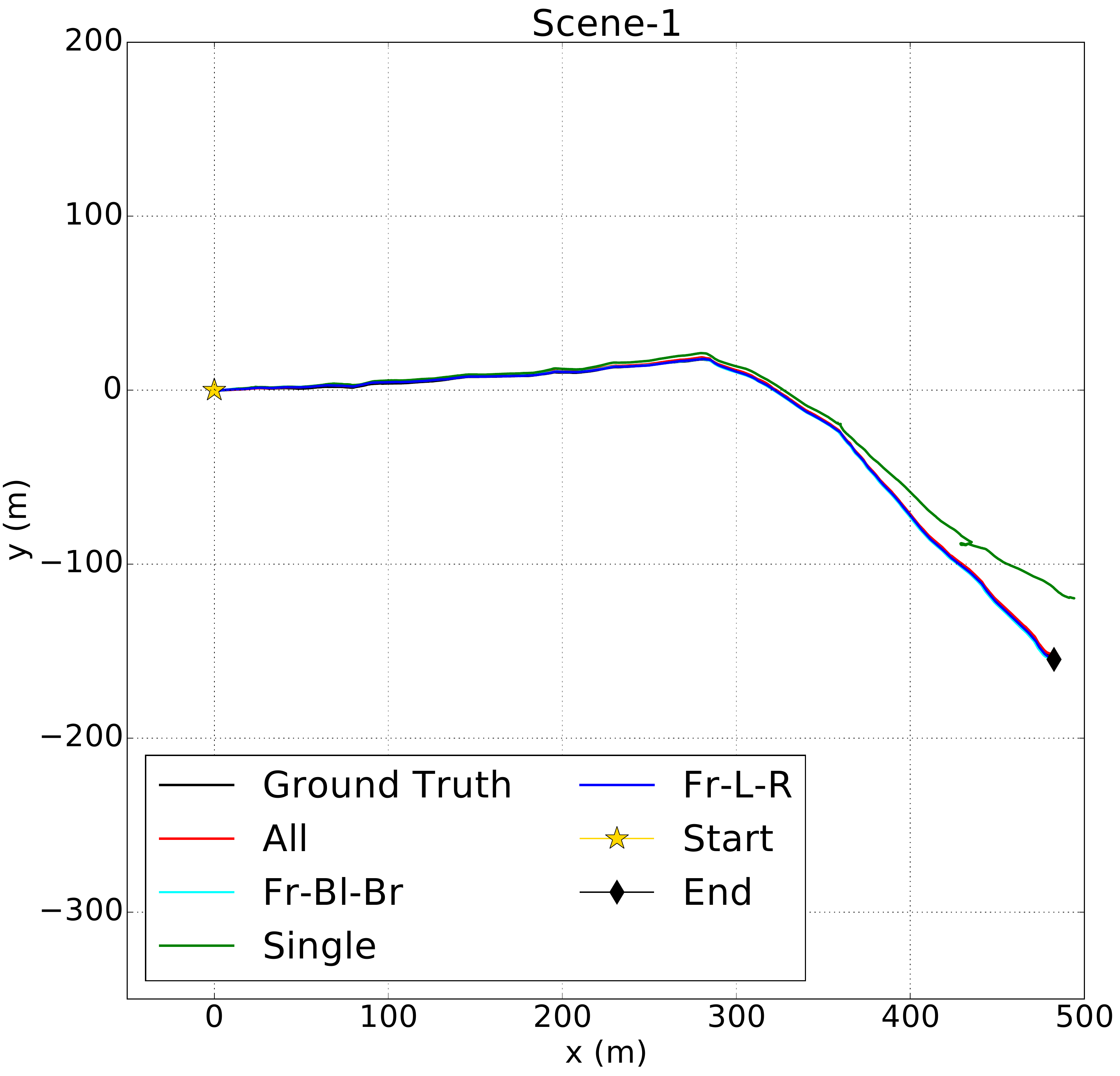}
			\includegraphics[width=0.495\columnwidth, height=4.5cm ]{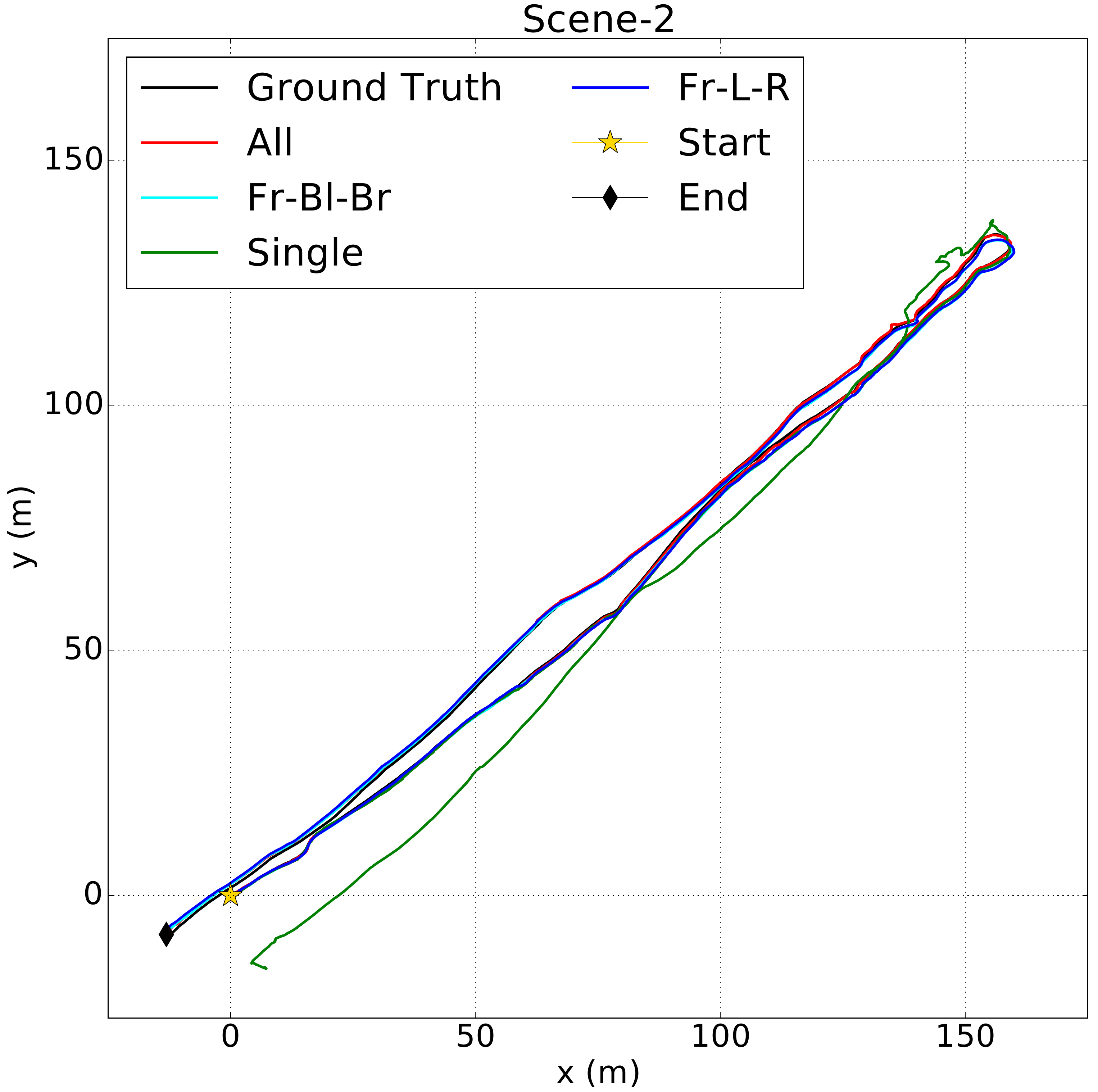}
			\caption{The comparison of trajectories generated (from the view of bird's eye, since the trajectories are very close with others, we strongly recommend the readers to zoom in vector graph for further details) from D-GPS and our algorithm with different LiDAR configurations. Where, \textit{GT} is the trajectory from D-GPS, \textit{All} is with all LiDAR enabled, \textit{Fr-Bl-Br} is the configuration with the front, back-left, back-right LiDARs enabled, \textit{Fr-L-R} is the configuration with the front, left, right LiDARs enabled, \textit{Single} is the configuration with only the front LiDAR enabled.}
			\label{fig_comp_trajecties}
		}
		\vspace{-0.5cm}
	\end{figure}

	While the proposed method converges qualitatively, in this sub-section, we perform quantitative evaluation on the localization accuracy of our algorithm by comparing with a differential Global Positioning System (D-GPS)\footnote{https://www.dji.com/d-rtk}, which can provide the localization reference with the precision of $1cm + 1ppm$. We evaluate our algorithm with different numbers of LiDARs on both of the two scenes. The comparison of different trajectories are shown in Fig. \ref{fig_comp_trajecties}, where the trajectory of single front LiDAR follows the Ground-Truth well at first but fails in long run due to the lack of sufficient features within the small FoV.
	
	Table~\ref{table:table_accuracy} shows the maximum \textit{absolute / relative error} among different configurations, showing that multiple LiDARs have great impact on improving the accuracy of localization. In addition, we plot the absolute error over distance in Fig. \ref{fig_absoluted_error} for detailed reference.
	
	In both scenes, we have achieved the precision of about $0.2\%$, which demonstrates that our algorithm is of high-accuracy.
	
	\subsection{Result of mapping}
	In the \textit{Scene-1}, the maps we built are shown in Fig. \ref{fig_maps}, with the point cloud data sampled from different LiDARs being rendered with different colors. From both the bird's eye-view (Fig. \ref{fig_maps}.A) and detailed view (Fig. \ref{fig_maps}.(C-E)), we can see that the point cloud data from different LiDARs is aligned well together and the consistency is kept both locally and globally. In summary, we have examined and verified the convergence and precision of the proposed algorithm on actual system with real world data.
	
	\begin{figure}[h]
		{
			\centering
			\includegraphics[width=1.00\columnwidth]{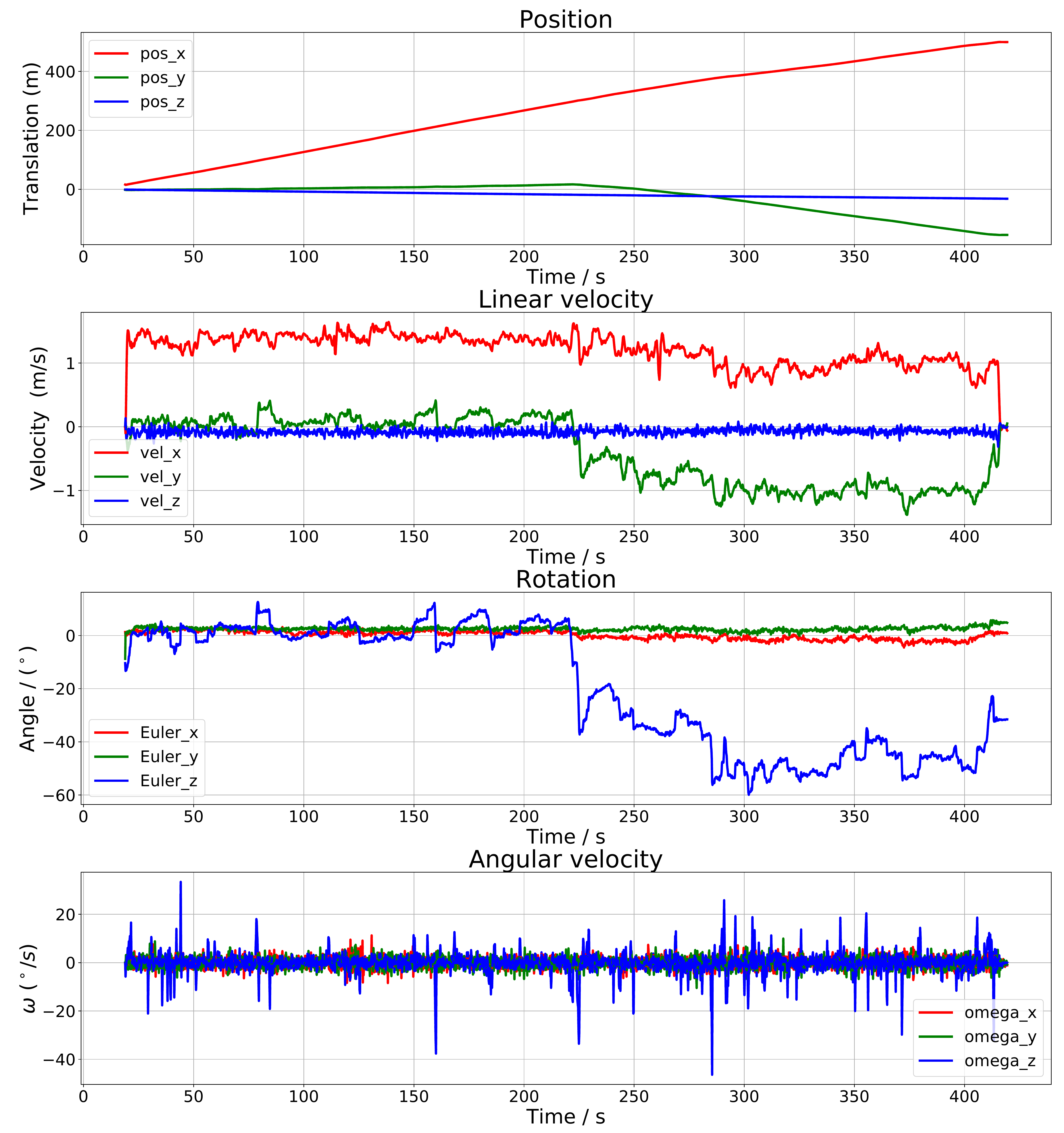}
			\caption{The estimation of position, linear velocity, rotation and angular velocity of geometric center in \textit{Scene-1}, with the configuration of \textit{All} LiDARs being enabled. The data plot starts after the map alignments, which is at $t=18.9s$.}
			\label{fig_state_esimation_scene_1}
		}
		\vspace{-0.5cm}
	\end{figure}
		
	\begin{table}[h]
		\setlength{\extrarowheight}{.1em}
		\setlength\extrarowheight{0.01pt}
		\begin{tabular}[h]{|c|c|c|c|c|}
			\hline
			& \scriptsize{All}& \scriptsize{\textit{Fr-Bl-Br}} &\scriptsize{\textit{Fr-L-R}}&\scriptsize{Single}\\
			& \scriptsize{Max (m / \%)} & \scriptsize{Max (m / \%)} & \scriptsize{Max (m / \%)} & \scriptsize{Max (m / \%)}\\
			\hline
			\textit{Scene-1}& \scriptsize{\textbf{1.17 / 0.21\%}} & \scriptsize{1.99 / 0.36\%} & \scriptsize{1.29 / 0.24\%} & \scriptsize{35.16 / 6.38\%} \\
			\hline
			\textit{Scene-2}& \scriptsize{\textbf{0.88 / 0.20\%}} & \scriptsize{1.19 / 0.27\%} & \scriptsize{1.33 / 0.31\%} & \scriptsize{14.90 / 3.41\%} \\
			\hline
		\end{tabular}
		\caption{The maximum absolute error (\textit{m}) and relative error (\textit{\%}) among different LiDAR configurations of \textit{Scene-1} and \textit{Scene-2}, whose total length are \textit{551.45m} and \textit{436.47m}, respectively.}
		\label{table:table_accuracy}
		\vspace{-0.5cm}
	\end{table}

	\begin{figure*}[htp]
		\centering
		\setcounter{figure}{10}
		\includegraphics[width=2.0\columnwidth]{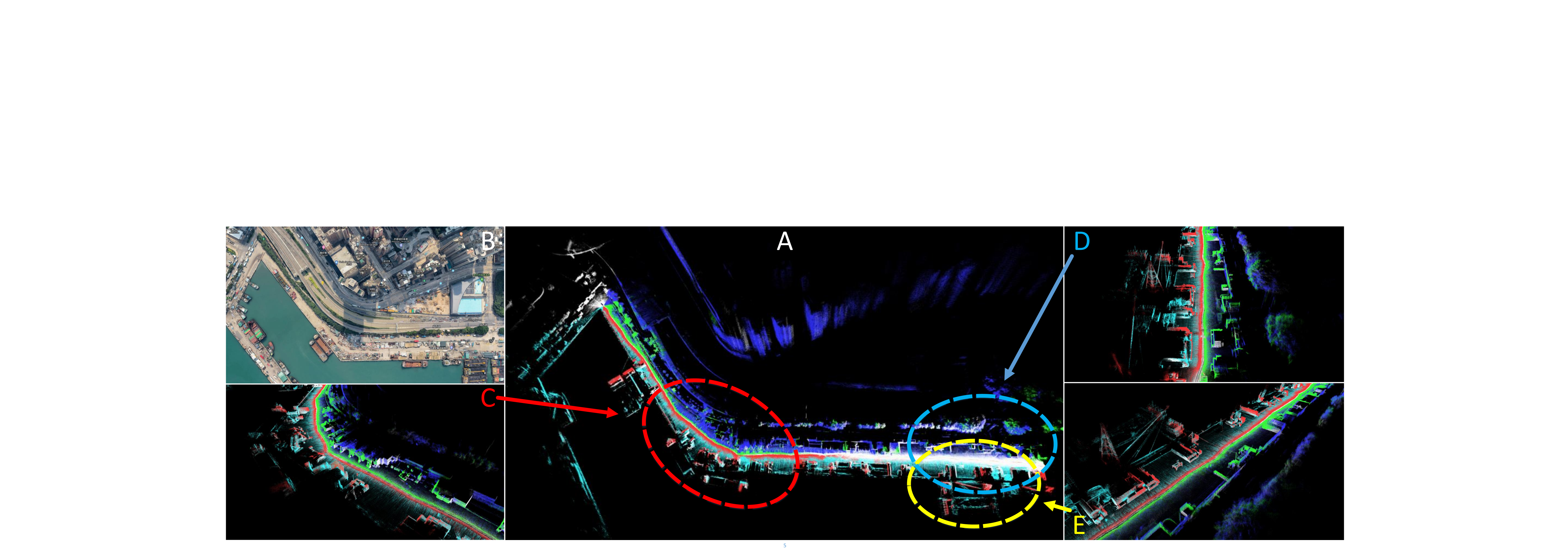}
		\caption{
			A): The bird's eye-view of the map we reconstruct with the data collected in \textit{Scene-1}. The point cloud data sampled from different LiDARs are rendered with different colors. The points of white, red, deep blue, cyan and green are the data sampled by the LiDAR installed on font, left, right, back-left and back-right, respectively; B): The satellite image of the experiment test ground; C)$\sim$E): The detailed inspection of the area marked in dashed circle in A.
		}
		\vspace{-0.5cm}
		\label{fig_maps}
	\end{figure*}
	
	\begin{figure}[H]
		{
			\setcounter{figure}{9}
			\includegraphics[width=0.49\columnwidth, height=4.0cm  ]{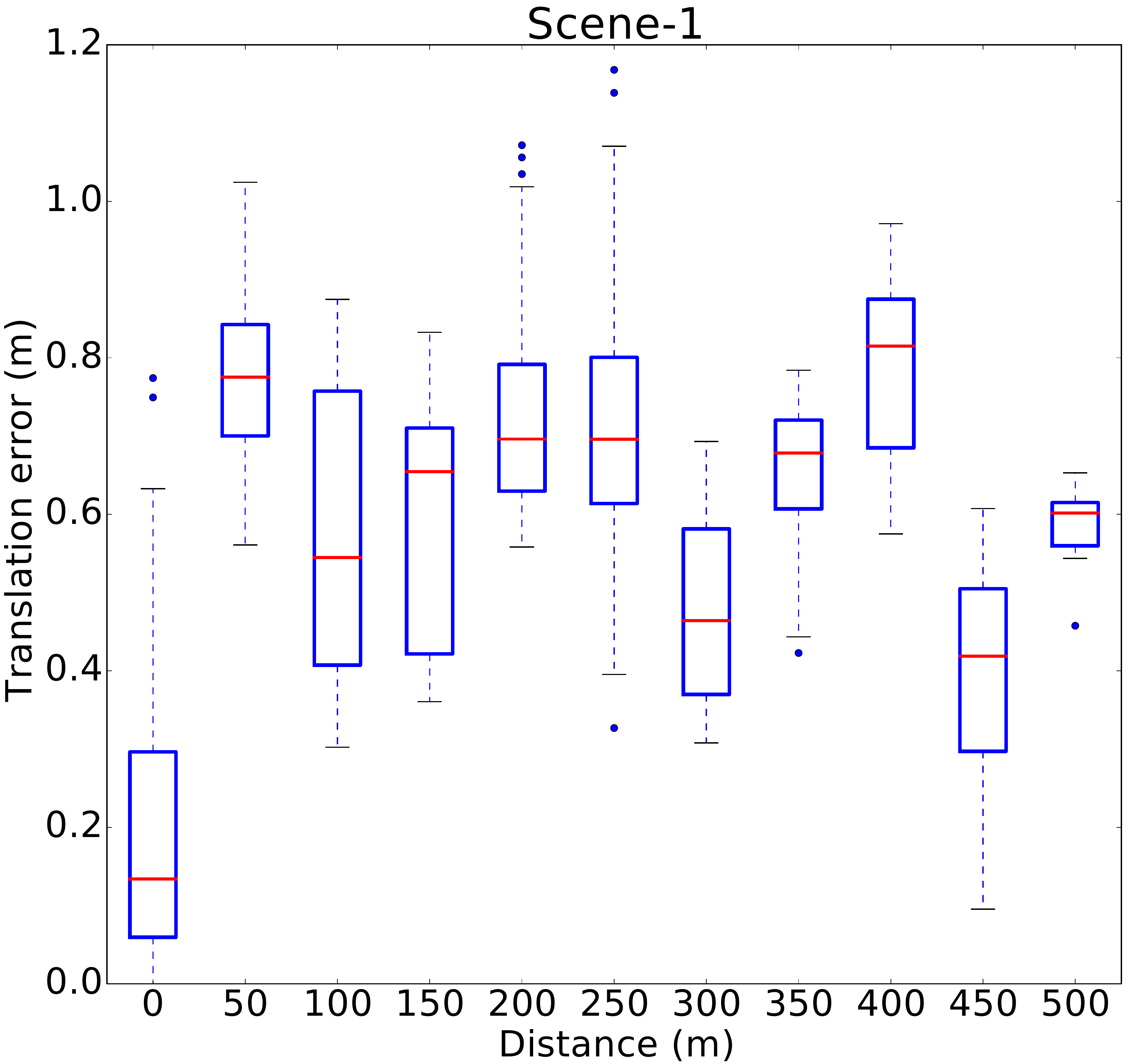}
			\includegraphics[width=0.49\columnwidth, height=4.0cm ]{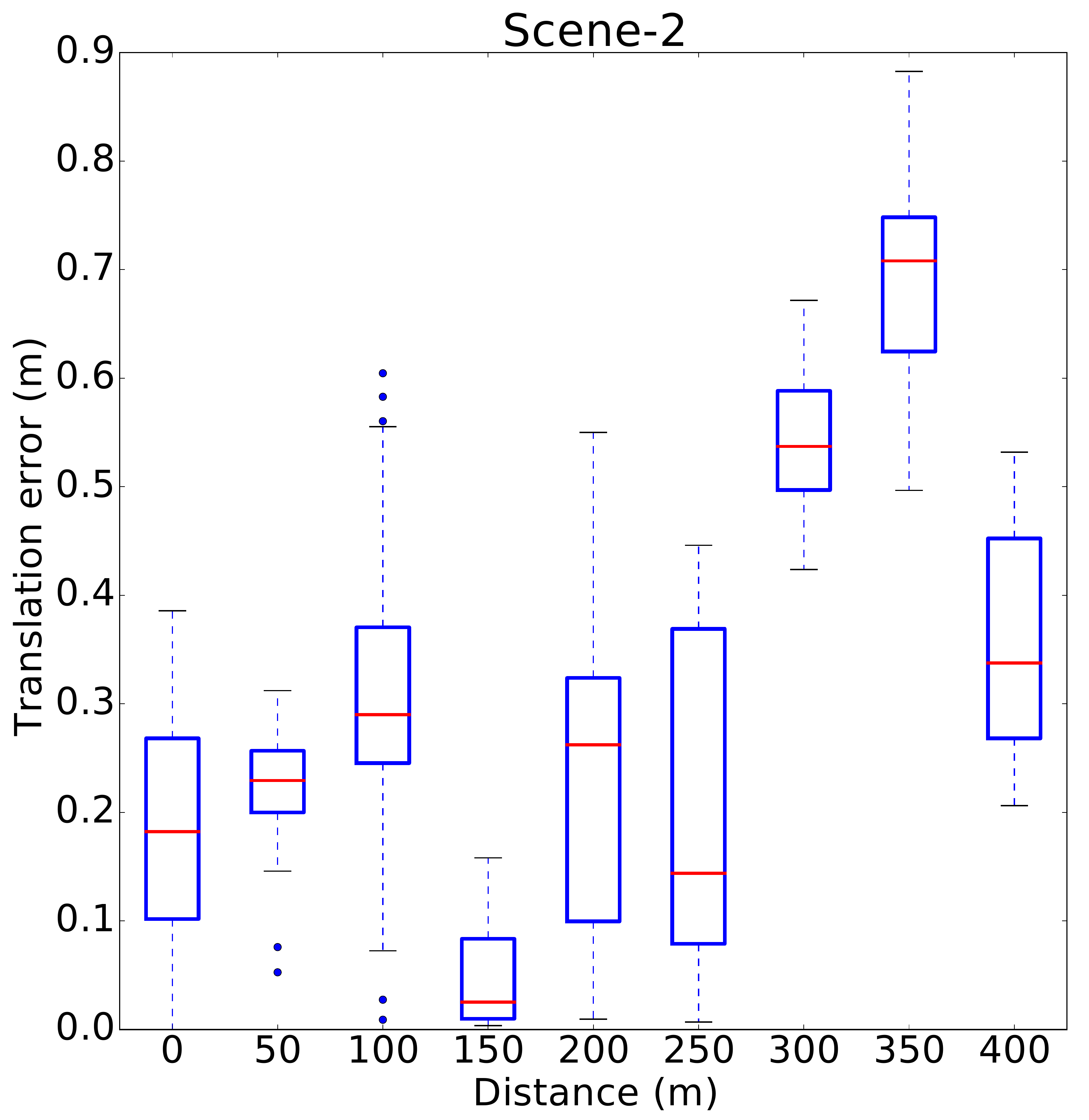}
			\caption{The absolute translation error over time, with the configuration of all LiDAR being enabled (our best accuracy).}
			\label{fig_absoluted_error}
		}
		\vspace{-0.3cm}
	\end{figure}
	
	\section{Conclusion and discussion}
	
	This paper presents a decentralized EKF algorithm for simultaneous calibration, localization, and mapping with multiple LiDARs. Experiments in urban area are conducted. Results show that the proposed algorithm converges stably and has achieved 0.2\% accuracy at low speed motion.
	
	As mentioned in our previous Section \ref{sect_experiments}, our current implementation is based on a single high-performance PC where all communication are done within a PC, problems such as message synchronization or communication loss do not occur and are not considered. Moreover, limited by the computing power of the PC, the current implementation runs offline. Future work will implement on each LiDAR dedicated computer, solve the problem therein (e.g., time synchronization) and verify its robustness in presence of LiDAR failure.
	
	\section{Acknowledgement}
	The authors would like to thank DJI Co., Ltd\footnote{\url{https://www.dji.com}}. for donating devices and research fund.
	
	\bibliography{iros2020jiarong}
	
\end{document}